\def\year{2018}\relax
\documentclass[letterpaper]{article} 
\usepackage{aaai18}  
\usepackage{times}  
\usepackage{helvet}  
\usepackage{courier}  
\usepackage{url}  
\usepackage{graphicx}  
\usepackage{amsmath}
\usepackage{amsfonts}
\usepackage{mathrsfs}
\usepackage{todonotes}
\usepackage{amsthm}
\usepackage[normalem]{ulem}

\presetkeys{todonotes}{inline}{}

\newtheorem{theorem}{Theorem}
\newtheorem{lemma}[theorem]{Lemma}
\newtheorem{assumption}[theorem]{Assumption}
\newtheorem{condition}[theorem]{Condition}
\newtheorem{definition}[theorem]{Definition}

\newtheorem{remark}[theorem]{Remark}
\newtheorem{corollary}[theorem]{Corollary}

\usepackage{paralist}

\usepackage{times}
\usepackage{graphicx} 
\usepackage{subfigure} 

\usepackage{algorithm}
\usepackage{algorithmic}

\usepackage{array}
\usepackage{natbib}
\usepackage{hyperref}

\def \E {\mathbb{E}}
\def \Pr {\mathrm{Pr}}

\def \vec {\mathrm{vec}}

\def \efs {\eta_{\delta,N}}

\def \stat {{\psi}}
\def \fstat {{\xi}}
\def \pstat {{\psi}^h}

\newcommand{\h}[1]{{\cal #1}}

\newcommand{\condop}[3]{V_{#1 \mid #2; #3}}
\def \ido {{\mathbf{do}}}

\def \Wsys {W_\mathrm{sys}}
\def \hWsys {\hat{W}_\mathrm{sys}}

\begin{document}

\title{An Efficient, Expressive and Local Minima-free Method for Learning Controlled Dynamical Systems}
\author{
Ahmed Hefny\\
Carnegie Mellon University
\And
Carlton Downey\\
Carnegie Mellon University
\And
Geoffrey Gordon\\
Carnegie Mellon University
}

\maketitle

\begin{abstract}
We propose a framework for modeling and estimating the state of controlled 
dynamical systems, where an agent can affect the system through actions
and receives partial observations.
Based on this framework, we propose the Predictive State Representation with Random Fourier Features (RFF-PSR).
A key property in RFF-PSRs is that the state estimate is represented by a conditional distribution of future observations given future actions.
RFF-PSRs combine this representation with moment-matching, kernel embedding and local optimization to
achieve a method that enjoys several favorable qualities: It can represent controlled environments which can be affected by actions;
it has an efficient and theoretically justified learning algorithm;
it uses a non-parametric representation that has expressive power to represent continuous non-linear dynamics.
We provide a detailed formulation, a theoretical analysis 
and an experimental evaluation that demonstrates the effectiveness of our method.

\end{abstract}

\section{Introduction}
\begin{table*}
\centering
{\small 
\begin{tabular}{|c|c|c|>{\centering}p{1cm}|>{\centering}p{1.6cm}|c|c|}
\hline
Method & Actions & Continuous & Non-linear & Partially observable & Scalable & Consistent
 \\
\hline
Non-linear ARX & \checkmark & \checkmark & \checkmark & $\times$ & \checkmark & \checkmark \\ 
N4SID for Kalman Filter & \checkmark & \checkmark & $\times$ & \checkmark & \checkmark & \checkmark \\
Non-convex optimization (e.g. EM) & \checkmark & \checkmark & \checkmark & \checkmark & \checkmark & $\times$ \\
Gram-Matrix (e.g. HSE-PSR) & \checkmark & \checkmark & \checkmark & \checkmark & $\times$ & \checkmark \\
Spectral PSR/POMDP & \checkmark & $\times$ & \checkmark & \checkmark & \checkmark &  \checkmark \\
Reduction to Supervised Learning & $\times$ & \checkmark & \checkmark & \checkmark & \checkmark & \checkmark \\
\hline
\textbf{RFF-PSR} & \checkmark & \checkmark & \checkmark & \checkmark & \checkmark & \checkmark \\
\hline
\end{tabular}
}
\caption{Comparison between proposed RFF-PSR and existing system identification methods in terms of the type of systems they can model
as well as their computational efficiency and statistical consistency.
The table should be interpreted as follows: for each method there exists
an instantiation that {\em simultaneously} satisfies all 
properties marked with $\checkmark$ but there is no instantiation 
that is guaranteed to satisfy the properties marked with $\times$.
A method is scalable if computational and memory costs scale at most linearly with the number of training examples.
For RFF-based methods, consistency is up to an approximation error that is controllable by the number of features~\citep{rff}.
}
\label{tbl:models}
\end{table*}

Controlled dynamical systems, where an agent can influence an 
environment through actions and receive partial observations, emerge
in numerous applications in robotics and automatic control.
Modeling and learning these systems from data is of great importance
in these fields.

The general problem of learning dynamical systems from data (also known as system identification) has been extensively studied and several methods were proposed to tackle it. However, having an expressive, efficient and consistent method for non-linear controlled systems remains an open problem.

Many system identification methods rely on likelihood-based optimization or sampling using EM, MCMC or gradient descent.
which makes them prone to poor local optima. There is another class of methods that alleviates the local optima problem and offers 
a tractable and statistically consistent approach to system identification.
These methods, usually referred to as spectral algorithms, have two key properties in common: predictive representation and method of moments.
Instead of the state being a latent variable, they represent the estimated state by the expectation of sufficient statistics (or features) of future observations; and they use method of moments to learn model parameters from data.\footnote{There is a class of spectral algorithms that maintains the latent variable view. This is exemplified
by tensor decomposition methods~\citep{tensor_factorization}.}

Initially introduced for linear-Gaussian systems~\citep{vanoverschee:96}, these algorithms have been extended to discrete systems~\citep{hkz,rrhmm,boots:11:plan} and then to general smooth continuous systems~\citep{hsepsr}.
More recently, it has been shown that a wide class of spectral learning algorithms 
for uncontrolled systems are instances of a two-stage regression framework~\citep{hefny:15},
where system identification is reduced to solving a set of regression problems.
This framework allows for seamless integration of compressing non-linearities, sparsity~\citep{xia2016}
and online learning~\citep{onlineivr} into system identification, and for establishing theoretical guarantees by leveraging 
the rich literature on supervised regression.

Unfortunately, the formulation in~\citep{hefny:15} is limited to uncontrolled systems.
On the contrary, we are interested in controlled systems, where the user can affect the system through actions.
This gives rise to a key issue: the policy that determines the actions can change at test time.
For this reason, the representation of the predictive state must be independent of the training policy
and therefore must encode a \emph{conditional distribution} of future observations given future actions. 
To adopt such a representation into a practical method 
that retains the benefits of the two-stage regression formulation,
there are a number of challenges that need to be tackled.

First, we need a suitable state representation and dynamics model that can be used to represent a wide class of 
controlled dynamical systems while ensuring the learning problem remains tractable.
Second, we would like to benefit from the two-stage regression view of~\citep{hefny:15}
to facilitate model formulation. 
However, a key assumption in that work is that future observations provide an unbiased estimate of the predictive state,
which is not true when the state is a conditional distribution. 
Third, having a different state representation and having action policy playing a key role on determining the training data
require a different theoretical analysis than the one in~\citep{hefny:15}.
Fourth, because they are based on method of moments, two stage regression models 
are statistically inefficient. Having the ability to refine the model using local optimization 
can lead to significant gains in predictive performance.

In this work we address these challenges by combining ideas from two-stage regression, kernel embedding and approximation and gradient descent with backpropagation through time to develop RFF-PSRs.
Overall, RFF-PSRs enjoy a number of advantages that, to our knowledge, are not attained by existing system identification methods. We
 summarize these advantages in Table \ref{tbl:models}.

In summary, the contributions of this work are as follows:
(1) We develop a two-stage regression framework for controlled dynamical systems that admits tractable learning (Sections \ref{sec:formulation}-\ref{sec:learning}). 
(2) Through the two-stage regression view, we provide theoretical guarantees on learning the parameters of a controlled system (Section  \ref{sec:theory}).
(3) We use the extended formulation to construct RFF-PSRs, 
an efficient approximation of kernel-based predictive state representations (HSE-PSRs) (Section \ref{sec:rffpsr}).
(4) We provide a means to refine the parameters of a controlled dynamical system
and apply it to our proposed RFF-PSR model (Section \ref{sec:refine}).
(5) We demonstrate the advantages of our proposed method through synthetic and robot simulation experiments (Section \ref{sec:exp}).

\section{Related Work}
Developing tractable and consistent algorithms for latent state dynamical systems dates back to spectral subspace identification algorithms 
for Kalman filters\citep{vanoverschee:96}. At their heart, these algorithms represent the state as a prediction of the future observations conditioned on history and future actions, and use matrix factorization to obtain a basis for the state.

This notion of the state as a prediction is the basis of \emph{predictive state representations} (PSRs)~\citep{psr}, where the state is represented by the 
success probabilities of a number of {\em tests}. A test succeeds if a specified sequence of test observations is observed when administering a specified sequence of test actions.

Noting that the state and parameters of a PSR are defined up to a similarity transformation has led to a family of tractable and consistent spectral algorithms 
for learning PSRs~\citep{tpsr}. 
More recently,~\citet{hsepsr} proposed a generalization of PSRs in a reproducing kernel Hilbert space (RKHS). 
This Hilbert space embedding of PSRs (HSE-PSRs) is able to represent systems with continuous
observations and actions while still offering a tractable and consistent learning algorithm. HSE-PSRs, however, use a Gram matrix formulation, whose computational
and storage requirements can grow rapidly with the size of training data.
A finite dimensional approximation for non-linear PSRs was proposed by~\citet{onlinepsr}.
However, it can be thought of as an approximation of HSE-HMMs~\citep{hsehmm}
with actions, a method that has poor theoretical guarantees~\citep{hsepsr}. 
In addition,~\citet{onlinepsr} did not provide examples of how to apply the proposed model to 
controlled processes with continuous actions. In contrast, the model we propose is an approximation of HSE-PSRs,
which is a more principled generalization of PSRs as it performs true Bayesian inference in the RKHS.
In addition, our proposed learning algorithm incorporates a local optimization procedure that we demonstrate to be very effective.

%
%

We use a reduction of system identification to supervised regression.
Similar reductions has been proposed in the literature~\citep{spr,hefny:15,onlinepsr,onlineivr,psim}.
These reductions, however, assume uncontrolled systems, where future observation 
statistics constitute an unbiased representation of the predictive state.\footnote{implicit reductions do exist in the system identification literature~\citep{vanoverschee:96} but they assume linear systems.}
Modeling controlled systems is more subtle since the 
the state of the system is a \emph{conditional distribution} of observations given actions.


Another related work is the spectral learning 
algorithm for POMDPs proposed by~\citet{specpomdp}. 
This method uses tensor factorization 
to recover POMDP parameters from examples collected by a non-blind memoryless policy. 
However, this method is limited to discrete POMDPs.
Also, PSRs have more representational capacity than POMDPs and can compactly represent more sophisticated systems~\citep{psr}. 
There are other classes of dynamical system learning algorithms
that are based on local optimization or sampling approaches~\citep{fox:09,gpss}
but they do not offer consistency guarantees.

\section{Formulation}
\label{sec:formulation}
We define a class of models that extends predictive state models of \citet{hefny:15} to controlled systems.
We first introduce some notation: 
We denote by $\Pr[x \mid \ido(Y=y)]$ the probability of $x$ given that we \emph{intervene} by setting $Y$ to $y$.
This is different from $\Pr[x \mid Y=y]$ which denotes conditioning on \emph{observing} $Y=y$; in the former case, we ignore all effects on $Y$ by other variables.
We denote by $\condop{A}{B}{c}$ the linear operator that satisfies
\begin{align*}
\E[A|B=b,C=c] = \condop{A}{B}{c} b \quad \forall b, c
\end{align*}
In other words for each $c$, $\condop{A}{B}{c}$ is a conditional expectation operator
from $B$ to $A$. In the discrete case, $\condop{A}{B}{c}$ is just a conditional probability table.

When dealing with multiple variables, we will use tensor notation; e.g.,
$V_{A,B \mid C,D}$ is a 4-mode tensor. We will use
\begin{align*}
V_{A,B \mid C,D} \times_C c \times_D d
\end{align*}
to denote multiplying $V_{A,B \mid C,D}$ by $c$ along the mode corresponding to $C$
and by $d$ along the mode corresponding to $D$. If $c$ is a matrix then the multiplication is performed along the first dimension of $c$.

We will also use $\| \cdot \|_F$ to denote Frobenius norm, $a \otimes b$ to denote Kronecker product of two vectors
and $A \star B$ to denote the Khatri-Rao product of two matrices (columnwise Kronecker product).

\subsection{Model Definition}
We will consider $k$-observable systems, 
where the posterior belief state given all previous observations and actions is uniquely identified by the conditional distribution
$\Pr[o_{t:t+k-1} \mid \ido(a_{t:t+k-1})]$.

Following \citet{hefny:15},
we denote by $\stat^o_t$, $\stat^a_t$, $\fstat^o_t$ and $\fstat^a_t$
sufficient features of future observations $o_{t:t+k-1}$, future actions
$a_{t:t+k-1}$, extended future observations $o_{t:t+k}$ 
and extended future actions $a_{t:t+k}$ at time $t$ respectively.

We also use $h^\infty_t \equiv o_{1:t-1},a_{1:t-1}$ to denote the entire history of observations and actions at time $t$ and use $\pstat_t \equiv \pstat(o_{1:t-1},a_{1:t-1})$  to denote finite features of previous observations
and actions before time $t$.\footnote{Often but not always, $\pstat_t$ is a computed from fixed-size window of previous observations and actions ending at $t-1$.}
 
We are now ready to define the class of systems we are interested in.
\begin{definition}
A dynamical system is said to conform to a \textbf{predictive state controlled model (PSCM)}
if it satisfies the following properties:
\begin{itemize}
\item For each time $t$, there exists a linear operator $Q_t = \condop{\stat^o_t}{\ido(\stat^a_t)}{h^\infty_t}$ (referred to as predictive state) such that
$\E[\stat^o_t \mid \ido(a_{t:t+k-1}), h^\infty_t] = Q_t \stat^a_t$
\item For each time $t$, there exists a linear operator $P_t = \condop{\fstat^o_t}{\ido(\fstat^a_t)}{h^\infty_t}$ (referred to as extended state) such that
$\E[\fstat^o_t \mid \ido(a_{t:t+k}), h^\infty_t] = P_t \fstat^a_t$
\item There exists a linear map $\Wsys$ (referred to as system parameter map), such that, for each time $t$, 
\begin{align}
P_t = \Wsys (Q_t)
\label{eq:form}
\end{align}
\item There exists a filtering function $f_\mathrm{filter}$ such that, for each time  $t$, $Q_{t+1} = f_\mathrm{filter}(P_t, o_t, a_t)$.
$f_\mathrm{filter}$ is typically non-linear but known in advance.
\end{itemize}
\end{definition}

It follows that a PSCM is specified by the tuple $(Q_0, \Wsys, f_\mathrm{filter})$, where $Q_0$ denotes the initial belief state.

There are a number of aspects of PSCMs that warrant discussion. 
First, unlike latent state models, the state $Q_t$ is represented by a conditional distribution of observed quantities.
Second, $Q_t$ 
is a deterministic function of the 
history $h^\infty_t$. It represents the \emph{belief} state 
that one should maintain after observing the history to make optimal predictions.
Third, a PSCM specifies a recursive filter where given an action $a_t$ and
an observation $o_t$, the state update equation is given by
\begin{align}
\label{eq:state_update}
Q_{t+1} = f_\mathrm{filter}(\Wsys(Q_t),o_t,a_t) 
\end{align}
This construction allows us to have a linear map $\Wsys$ and still use it
to build models with non-linear state updates,
including IO-HMMs~\citep{iohmm}, Kalman filters with inputs~\citep{vanoverschee:96}
and HSE-PSRs~\citep{hsepsr}.
As we see in Section \ref{sec:learning}, avoiding latent variables and having a linear 
$\Wsys$ enable the formulation of a consistent learning algorithm.



\section{Learning A Predictive State Controlled Model}
\label{sec:learning}
We assume that the extended features $\fstat^o_t$ and $\fstat^a_t$ are chosen such that
$f_\mathrm{filter}$ is known. The parameters to learn are thus $\Wsys$ and $Q_0$.
We also assume that a fixed blind (open-loop) policy is used to collect training data, 
and so we can treat causal conditioning on action $\ido(a_t)$ as ordinary conditioning on $a_t$.\footnote{One way to deal with non-blind training policies is to assign importance weights to training examples to
correct the bias resulting from non-blindness~\citep{nblindpsr,boots:11:plan}. This, however, requires knowledge of the data collection
policy and can result in a high variance of the estimated parameters. We defer the case of unknown non-blind
policy to future work.}
It is possible, however, that a different (possibly non-blind) policy is used at test time.

To learn model parameters, we will adapt the two-stage regression method of \citet{hefny:15}.
Let $\bar{Q}_t \equiv \E[Q_t \mid \pstat_t]$ 
(resp. $\bar{P}_t \equiv \E[P_t \mid \pstat_t]$)
be the expected state (resp. expected extended state)
conditioned on finite history features $\pstat_t$.
For brevity, we might refer to $\bar{Q}_t$ simply as the (predictive) state
when the distinction from $Q_t$ is clear.
It follows from linearity of expectation that
$\E[\stat^o_t \mid \stat^a_t, \pstat_t] = \bar{Q}_t \stat^a_t$
and $\E[\fstat^o_t \mid \fstat^a_t, \pstat_t] = \bar{P}_t \fstat^a_t$;
and it follows from the linearity of $\Wsys$ that
\begin{align*}
\bar{P}_t = \Wsys (\bar{Q}_t)
\end{align*}

So, we train regression models (referred to S1 regression models) to
estimate $\bar{Q}_t$ and $\bar{P}_t$ from $\pstat_t$. 
Then, we train another (S2) regression model to estimate $\Wsys$
from $\bar{Q}_t$ and $\bar{P}_t$. Being conditional distributions, estimating $\bar{Q}_t$ and $\bar{P}_t$ from $\pstat_t$ is more subtle compared to uncontrolled systems, since we cannot use
observation features as estimates of the state.
We describe two methods to construct an S1 regression model to estimate $\bar{Q}_t$. The same methods apply to $\bar{P}_t$.
As we show below, instances of both methods exist in the literature of system identification.

\subsection{Joint S1 Approach}
\label{sec:joint}
Let $\stat^{oa}_t$ denote a sufficient statistic of the joint observation/action distribution 
$\Pr(\stat^o_t, \stat^a_t \mid \pstat_t)$. This distribution is fixed for each value of $\pstat_t$ since we assume a fixed model and policy.  
We use an S1 regression model to learn the map $f : \pstat_t \mapsto \E[\stat^{ao}_t \mid \pstat]$ by solving the optimization problem
\begin{align*}
\arg \min_{f \in {\cal F}} \sum_{t=1}^T l(f(\pstat_t), \stat^{oa}_t) + R(f)
\end{align*}
for some suitable Bregman divergence loss $l$ (e.g., square loss)
and regularization $R$.

Once we learn $f$, we can estimate $\bar{Q}_t$ by first estimating the joint distribution
$\Pr(\stat^o_t, \stat^a_t \mid \pstat_t)$ and then deriving the conditional operator $\bar{Q}_t$. By the continuous mapping theorem, a consistent estimator of $f$ results in
a consistent estimator of $\bar{Q}_t$.
An example of applying this method is using kernel Bayes rule~\citep{kbr} to estimate states in HSE-PSR~\citep{hsepsr}.

\subsection{Conditional S1 Approach}
\label{sec:perror}
In this method, instead of estimating the joint distribution represented by $\E[\stat^{oa}_t \mid \pstat_t]$, we directly estimate the conditional
distribution $\bar{Q}_t$. We exploit the fact that each training example 
$\stat^o_t$ is an unbiased estimate of $\bar{Q}_t \stat^a_t = \E[\stat^o_t \mid \stat^a_t, \pstat_t]$.
We can formulate the S1 regression problem as
learning a function $f : \pstat_t \mapsto \bar{Q}_t$ that best matches the training examples, i.e., we solve the problem
\begin{align}
\arg \min_{f \in {\cal F}} \sum_{t=1}^T l(f(\pstat_t)\stat^a_t, \stat^o_t) + R(f)
\label{eq:perror}
\end{align}
for some suitable Bregman divergence loss $l$ (e.g., square loss) and regularization $R$.
An example of applying this method is the oblique projection method
used in spectral system identification~\citep{vanoverschee:96}. It is worth emphasizing that both the joint and conditional S1 approaches
assume the state to be a \emph{conditional} distribution. 
They only differ in the way to estimate that distribution.

\subsection{S2 Regression and Learning Algorithm}
\label{sec:s2reg}
Given S1 regression models to estimate $\bar{Q}_t$ and $\bar{P}_t$,
learning a controlled dynamical system proceeds as shown in Algorithm 
\ref{alg:learn}.

\begin{algorithm}
\begin{algorithmic}
\STATE \textbf{Input:} $\pstat_{n,t}$,$\stat^o_{n,t}$, $\stat^a_{n,t}$, $\fstat^o_{n,t}$, $\fstat^a_{n,t}$ for $1 \leq n \leq N$, $1 \leq t \leq T_n$ ($N$ is the number of trajectories, $T_n$ is the length of $n^{th}$ trajectory)
\STATE \textbf{Output:} Dynamics matrix $\hWsys$ and initial state $\hat{Q}_0$
\STATE Use S1A regression to estimate $\bar{Q}_{n,t}$.
\STATE Use S1B regression to estimate $\bar{P}_{n,t}$.
\STATE Let $\hWsys$ be the (regularized) least squares solution to the system of equations 
\begin{align*}
\bar{P}_{n,t} \approx \Wsys (\bar{Q}_{n,t}) \quad \forall n,t
\end{align*}
\IF{$N$ is sufficiently large}
\STATE Let $\bar{Q}_0$ be the (regularized) least square solution to the system of equations 
$\stat^o_{n,1} \approx Q_0 \stat^a_{n,1} \quad \forall n$
\ELSE
\STATE
Set $\hat{Q}_0$ to the average of $\bar{Q}_{n,t}$
\ENDIF

\end{algorithmic}

\caption{Two-stage regression for predictive state controlled models}
\label{alg:learn}
\end{algorithm}

\subsection{Theoretical Guarantees}
\label{sec:theory}
It is worth noting that Algorithm \ref{alg:learn} is still an instance of the two stage regression framework described in~\citep{hefny:15} and hence retains its theoretical guarantees: mainly that we can bound the error in estimating the dynamics matrix $\Wsys$ in terms of S1 regression error bounds, assuming that we collect examples from the 
stationary distribution of a blind policy with sufficient exploration.

A blind policy provides sufficient exploration if it has a stationary distribution that (1)
visits a sufficient history set such that the set of equations $\E[P_t | \pstat_t] = W_{sys} (\E[Q_t | \pstat_t])$ are sufficient for estimating $W_{sys}$ and 
(2) provides training data to estimate $\E[Q_t | \pstat_t]$ and $\E[P_t | \pstat_t]$ with increasing accuracy.

\begin{theorem}
Let $\pi$ be a blind data collection policy with a stationary distribution. 
If history, action and observation features have bounded norms,
$\pi$ provides sufficient exploration, and ridge regression is used with $\lambda_1$ and $\lambda_2$ regularization parameter for S1 and S2 regression respectively, then
for all valid states $Q$ the following is satisfied with probability at least $1-\delta$.

\begin{align*}
& \| (\hat{W}_\mathrm{sys} - \Wsys) (Q) \| \leq \\
& O\left(\efs \left((1/\lambda_2) + (1/\lambda_2^\frac{3}{2}) \sqrt{1 + \sqrt{\frac{\log(1/\delta)}{N}}} \right)\right) \\
& + O\left(\frac{\log(1/\delta)}{\sqrt{N}} \left(\frac{1}{\lambda_2} + \frac{1}{\lambda_2^\frac{3}{2}} \right)\right) 
 + O\left( \sqrt{\lambda_2} \right),
\end{align*}
where 
\begin{align*}
    \efs = O_p\left(\frac{1/\sqrt{N} + \lambda_1}{c + \lambda_1}\right), 
\end{align*}
where $c > 0$ is a problem-dependent constant.
\label{thm:main}
\end{theorem}
We provide proofs and discussion of sufficient exploration condition in the supplementary material.

\section{Predictive State Controlled Models With Random Fourier Features}
\label{sec:rffpsr}
Having a general framework for learning controlled dynamical systems,
we now focus on HSE-PSR~\citep{hsepsr} as a non-parametric instance of that framework using 
Hilbert space embedding of distributions.
We first describe HSE-PSR learning as a two-stage regression method.
Then we demonstrate how to obtain a finite 
dimensional approximation using random Fourier features (RFF)~\citep{rff}.
Before describing HSE-PSR we give some necessary background on
Hilbert space embedding and random Fourier features.

\subsection{Hilbert Space Embedding of Distributions}
We will briefly describe the concept of Hilbert space embedding of distributions. We refer the reader to~\citep{hse} for more details on this topic.
Hilbert space embedding of distributions provide a non-parametric generalizations
of marginal, joint and conditional probability tables of discrete variables to continuous domains: namely,
mean maps, covariance operators and conditional operators.

Let $k$ be a kernel associated with a feature map $\phi(x)$ such that $k(x_1,x_2) = \langle \phi(x_1), \phi(x_2) \rangle$.
A special case for discrete variables is the delta kernel where $\phi(x)$ maps $x$ to an indicator vector.
For a random variable $X$, the \emph{mean map} $\mu_X$ is defined as $\E[\phi_\h{X}(X)]$. Note that $\mu_X$ is an element of the reproducing kernel Hilbert space (RKHS)
associated with $k$.

The uncentered covariance operator of two variables $X$ and $Y$ is 
$C_{XY} = \E[\phi_\h{X}(X) \otimes \phi_\h{Y}(Y)]$.
For universal kernels $k_\h{X}$ and $k_\h{Y}$, $C_{XY}$ is a sufficient representation of the joint distribution $\Pr(X,Y)$.
In this paper, we will use $C_{XY \mid z}$ to denote the covariance of $X$ and $Y$ given that $Z = z$.

Under smoothness assumptions, \citep{song:09} show that
$V_{\phi_\h{X}(X) \mid \phi_\h{Y}(Y)} = C_{XY} C_{XX}^{-1},$
where the conditional operator $V$ is as defined in Section \ref{sec:formulation}.
More generally,
$\condop{\phi_\h{X}(X)}{\phi_\h{Y}(Y)}{z} = C_{XY \mid z} C_{XX \mid z}^{-1}$.

\subsection{HSE-PSR as a predictive state controlled model}
\label{sec:rffpsr_pscm}
HSE-PSR is a generalization of IO-HMM that has proven to be successful in practice~\citep{hsepsr,Boots13NIPSb}.
It is suitable for high dimensional and continuous observations and/or actions.
HSE-PSR uses kernel feature maps as sufficient statistics of observations and actions.
We define four kernels $k_O, k_A, k_o, k_a$ over future observation features, 
future action features, individual observations and individual actions respectively. 

We can then define $\stat^{o}_t = \phi_O(o_{t:t+k-1})$ and 
similarly $\stat^{a}_t = \phi_A(a_{t:t+k-1})$.
We will also use $\phi^o_t$ and $\phi^a_t$
as shorthands for $\phi_o(o_t)$ and $\phi_a(a_t)$.
The extended future is then defined as $\fstat^{o}_t = \stat^o_t \otimes \phi^o_t$ and 
$\fstat^{a}_t = \stat^a_t \otimes \phi^a_t$

Under the assumption of a blind learning policy, the operators $Q_t$ and $P_t$
are defined to be
\begin{align}
Q_t & = \condop{\stat^o_t}{\stat^a_t}{h^\infty_t} \\
P_t & = (P_t^\xi, P_t^o) = (\condop{\stat^o_{t+1} \otimes \phi^o_t}{\stat^a_{t+1} \otimes \phi^a_t}{h^\infty_t}, \condop{\phi^o_t \otimes \phi^o_t}{\phi^a_t}{h^\infty_t}) 
\end{align}

Therefore, $Q_t$ specifies the state of the system as a conditional distribution of future observations 
given future actions while $P_t$ is a tuple of two operators that allow us to condition on the pair $(a_t,o_t)$
to obtain $Q_{t+1}$. In more detail, filtering in an HSE-PSR is carried out as follows
\begin{itemize}
\item From $o_t$ and $a_t$, obtain $\phi^o_t$ and $\phi^a_t$.
\item Compute $C_{o_to_t \mid h^\infty_t,a_t} = \condop{\phi^o_t \otimes \phi^o_t}{\phi^a_t}{h^\infty_t} \phi^a_t$
\item Multiply by inverse observation covariance to change ``predicting $\phi^o_t$'' into ``conditioning on $\phi^o_t$'': 
\begin{flalign*}
&\condop{\stat^o_{t+1}}{\stat^a_{t+1},\phi^o_t,\phi^a_t}{h^\infty_t} \\
& \quad = \condop{\stat^o_{t+1} \otimes \phi^o_t}{\stat^a_{t+1},\phi^a_t}{h^\infty_t} \times_{\phi^o_t}  (C_{o_to_t \mid h^\infty_t,a_t} + \lambda I)^{-1} 
\end{flalign*}
\item Condition on $\phi^o_t$ and $\phi^a_t$ to obtain shifted state
\begin{align*}
Q_{t+1} & \equiv \condop{\stat^o_{t+1}}{\stat^a_{t+1}}{\phi^o_t,\phi^a_t,h^\infty_t} \\
& =\condop{\stat^o_{t+1}}{\stat^a_{t+1},\phi^o_t,\phi^a_t}{h^\infty_t} \times_{\phi^o_t} \phi^o_t \times _{\phi^a_t} \phi^a_t
\end{align*}
\end{itemize}

Thus, in HSE-PSR, the parameter $\Wsys$ is composed of two linear maps; $f_o$ and $f_\fstat$ such that 
$P_t^\fstat = f_\fstat(Q_t)$ and $P_t^o = f_o(Q_t)$.
In the following section we show how to estimate $\bar{Q}_t$
and $\bar{P}_t$ from data. Estimation of $f_\fstat$, $f_o$ can then be carried out using kernel regression.

Learning and filtering in an  HSE-PSR can be implicitly carried out in the RKHS using a Gram matrix formulation. We will describe learning in terms of the RKHS
elements and refer the reader to~\citep{hsepsr} for details on the Gram matrix formulation.
As we mention in Section \ref{sec:rffpsr}, random Fourier features, 
provides a scalable approximation to operating in the RKHS.

\subsection{S1 Regression for HSE-PSR}
\label{sec:rffpsr_s1}
As discussed in section \ref{sec:learning} we can use a joint or conditional approach for S1 regression. We now demonstrate how these two approaches apply to HSE-PSR.

\subsubsection{Joint S1 Regression for HSE-PSR}
\label{sec:joints1rff}
This is the method used in~\citep{hsepsr}. In this approach we exploit the fact that 
\begin{align*}
\bar{Q}_t = W_{\stat^o_t | \stat^a_t; \pstat_t} = C_{\stat^o_t \stat^a_t | \pstat_t} (C_{\stat^a_t \stat^a_t | \pstat_t} + \lambda I)^{-1}
\end{align*}
So, we learn two linear maps $T_{oa}$ and $T_a$ such that
$T_{oa} (\pstat_t) \approx C_{\stat^o_t \stat^a_t \mid \pstat_t}$ 
and $T_a (\pstat_t) \approx C_{\stat^a_t \stat^a_t \mid \pstat_t}$.
The training examples for $T_{oa}$ and $T_a$ consist of pairs
$(\pstat_t,\stat^o_t \otimes \stat^a_t)$ and $(\pstat_t,\stat^a_t \otimes \stat^a_t)$
respectively.

Once we learn this map, we can estimate $C_{\stat^o_t \stat^a_t | \pstat_t}$
and $C_{\stat^a_t \stat^a_t | \pstat_t}$ and consequently estimate 
$\bar{Q}_t$. 

\subsubsection{Conditional S1 Regression for HSE-PSR}
\label{sec:conds1rff}
It is also possible to apply the conditional S1 regression formulation in Section \ref{sec:perror}.
Specifically, let $\cal{F}$ be the set of 3-mode tensors, with modes corresponding
to $\stat^o_t$, $\stat^o_t$ and $\pstat_t$. We estimate a tensor $T^*$ by optimizing
\begin{align*}
T^* = \arg\min_{T \in \cal{F}} \| (T \times_{\pstat_t} \pstat_t  \times_{\stat_a^t} \stat_a^t) - \stat_o^t \|^2 + \lambda \|T\|_{HS}^2,
\end{align*}
where $\|.\|_{HS}^2$ is the Hilbert-Schmidt norm, which translates to Frobenius 
norm in finite-dimensional Euclidan spaces. 
We can then use
\begin{align*}
\bar{Q}_t = T^* \times_{\pstat_t} \pstat_t
\end{align*}

For both regression approaches, the same procedure can be used to estimate the extended state 
$\bar{P}_t$ by replacing features $\stat^o_t$ and  $\stat^a_t$ with their extended counterparts $\fstat^o_t$ and  $\fstat^a_t$. 

\subsection{Approximating HSE-PSR with Random Fourier Features}
A Gram matrix formulation of the HSE-PSR  
has computational and memory requirements that grow rapidly with the number of training examples.
To alleviate this problem, we resort to kernel approximation---that is,
we replace RKHS vectors such as $\psi_t^o$ and $\psi_t^a$ with finite dimensional vectors 
that approximately preserve inner products.
We use random Fourier features (RFF)~\citep{rff} as 
an approximation but it is possible to use other approximation methods.
Unfortunately RFF approximation can typically require $D$ to be prohibitively large.
Therefore, we apply principal component analysis (PCA) to the feature maps to reduce their dimension to $p \ll D$. 
We apply PCA again
to quantities that require $p^2$ space such as extended features $\fstat^o_t$, $\fstat^a_t$ 
and states $\bar{Q}_t$, reducing them to $p$ dimensions. 
We map them back to $p^2$ dimensions when needed (e.g., for filtering).
We also employ randomized SVD~\citep{halko:11:randsvd} for fast computation of PCA,
resulting in an algorithm that scales linearly with $N$ and $D$.
\footnotetext{We provide pseudo-code in the supplementary material. MATLAB source code is available at: \url{https://github.com/ahefnycmu/rffpsr}}

\subsection{Model refinement by local optimization}
\label{sec:refine}
A common practice is to use the output of a moment-based algorithm 
to initialize a non-convex optimization algorithm such as EM~\citep{belanger:15:textlds} or gradient descent~\citep{jiang:16:psrgd}. 
Since EM is not directly applicable to RFF-PSRs, we propose a
gradient descent approach.
We can observe that filtering in an RFF-PSR defines a recurrent structure given by.
\begin{align*}
q_{t+1} & = f_\mathrm{filter}(\Wsys q_t, o_t, a_t), \\
\E[o_t|q_t] & = W_\mathrm{pred} (q_t \otimes \phi(a_t)), 
\end{align*}
where $W_\mathrm{pred}$ is a linear operator that predicts the next observation.\footnote{The linearity of $W_\mathrm{pred}$ is a valid assumption for a universal kernel.}
If $f_\mathrm{filter}$ is differentiable, we can improve our estimates
of $\Wsys$ and $W_\mathrm{pred}$ using backpropagation through time (BPTT)~\citep{bptt}. 
In particular, we optimize the error in predicting (features of) a window of observations. In our experiments, we learn to predict $o_{t:t+k-1}$ given $a_{t:t+k-1}$.

\begin{figure*}
\centering
\includegraphics[scale=0.43]{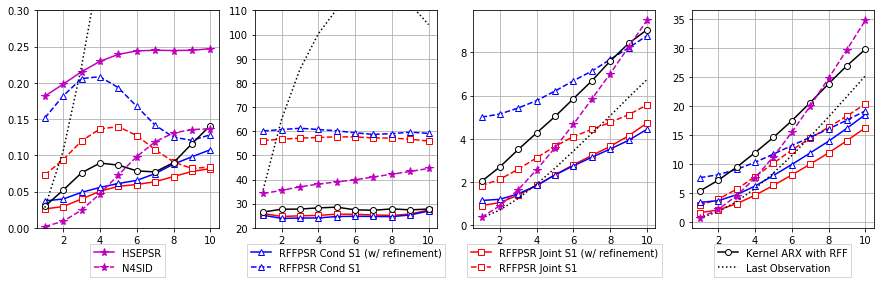}
\caption{Mean square error for 10-step prediction on (from left to right) synthetic model, 
TORCS car simulator, swimming robot simulation with 80\% blind test-policy, and swimming robot with 20\% blind test policy.
Randomly initialized RFF-PSR obtained significantly worse MSE and are not shown for clarity. A comparison with HSE-PSR on TORCS and swimmer datasets was not possible as it required prohibitively large memory.}
\label{fig:results}
\end{figure*}

\section{Experiments}
\label{sec:exp}

\subsection{Synthetic Data}
We use the benchmark synthetic non-linear system used by~\citep{hsepsr}
:
\begin{align*}
    \dot{x}_1(t) & = x_2(t) - 0.1 \cos (x_1(t))(5x_1(t)-4x_1^3(t)+x_1^5(t)) \\
    & \quad -0.5 \cos(x_1(t))a(t) \\
    \dot{x}_2(t) & = -65 x_1(t) + 50 x_1^3(t) - 15 x_1^5(t)-x_2(t)-100a(t) \\
    o(t) & = x_1(t)
\end{align*}
The input $a$ is generated as zero-order hold white noise, uniformly distributed between $-0.5$ and 0.5.
We collected 20 trajectories of 100 observations and actions at 20Hz and we split them into 10 training, 5 validation and 5 test trajectories. The prediction target for this experiment is $o(t)$.

\subsection{Predicting windshield view}
\begin{figure}[H]
\centering
\includegraphics[scale=0.3]{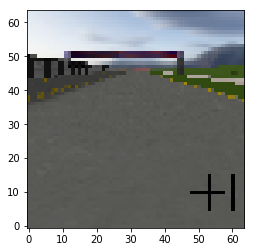} \\
\caption{An example of windshield view output by TORCS.}
\label{fig:torcs}
\end{figure}

In this experiment we used the TORCS car simulation server, which outputs 64x64 images 
(see Figure \ref{fig:torcs}). The observations are produced by converting the images to greyscale and
projecting them to 200 dimensions via PCA. The car is controlled by a built-in controller that controls acceleration
while the external actions control steering. We collected 50 trajectories by applying a sine wave with random starting phase to 
the steering control and letting the simulator run until the car goes off the track.
We used 40 trajectories for training, 5 for validation and 5 for testing. The prediction target is the projected image.

\subsection{Predicting the nose position of a simulated swimmer robot}
We consider the 3-link simulated swimmer robot from the open-source package RLPy~\citep{RLPy}. The 2-d action consists of torques applied on the two joints of the links. The observation model returns the angles of the joints and the position of the nose (in body coordinates).
The measurements are contaminated with Gaussian noise whose standard deviation is 5\% of the true signal standard deviation.
To collect the data, we use an open-loop policy that selects actions
uniformly at random. We collected 25 trajectories of length 100 each
and use 24 for training and 1 for validation. 
We generate test trajectories using a mixed policy: 
with probability $p_{\mathrm{blind}}$, we sample a uniformly random action, while with probability $1-p_{\mathrm{blind}}$, 
we sample an action from a pre-specified deterministic policy that seeks a goal point. 
We generate two sets of 10 test trajectories each, one with $p_{\mathrm{blind}} = 0.8$ and another with $p_{\mathrm{blind}} = 0.2$.
The prediction target is the position of the nose.
 

\subsection{Tested Methods and Evaluation Procedure}
We tested three different initializations of RFF-PSR (with Gaussian RBF kernel): random initialization, two-stage regression with joint S1, and two-stage regression with conditional S1 (Section \ref{sec:rffpsr_s1}). For each initialization, we tested the model before and after refinement. For refinement we used BPTT with a decreasing step size: the step size is reduced by half if validation error increases. Early stopping occurs if the step size becomes too small ($10^{-5}$) or the relative change in validation is insignificant ($10^{-3}$). We also test the following baselines.

\textbf{HSE-PSR}: We implemented the Gram matrix HSE-PSR
as described in \citep{hsepsr}. 

\textbf{N4SID}: We used MATLAB's implementation of subspace identification of linear dynamical systems. 

\textbf{Non-linear Auto Regression (RFF-ARX)}: 
We implemented a version of auto regression where the predictor variable is the RFF
representation of future actions together with a finite history of 
previous observations and actions, and the target variable is future observations. 

Models were trained with future length of 10
and history length of 20. For RFF-PSR and RFF-ARX we used 10000 random features and applied PCA to project features onto 20 dimensions. Kernel bandwidths were set to the median of the distance between training points (median trick). 
For evaluation, we perform filtering on the data and estimate 
the prediction target of the experiment at test time $t$ given the history $o_{1:t-H},a_{1:t}$,
where $H$ is the prediction horizon. We report the mean square error across
all times $t$ for each value of $H \in \{1,2,\dots,10\}$.

\subsection{Results and Discussion}
The results are shown in Figure \ref{fig:results}.
There are a number of important observations.

\begin{itemize}
\item In general, joint S1 training closely matches or outperforms conditional S1 training, with and without refinement.
\item Local refinement significantly improves predictive 
performance for all initialization methods.
\item Local refinement, on its own, is not sufficient to produce a 
good model. The two stage regression provides a good initialization
of the refinement procedure.
\item Even without refinement, RFF-PSR outperforms HSE-PSR.
This could be attributed to the dimensionality reduction step, which adds appropriate inductive bias.
\item Compared to other methods, RFF-PSR has better performance with non-blind test policies.
\end{itemize}

\section{Conclusion}
We proposed a framework to learn controlled dynamical systems 
using two-stage regression. We then applied this framework to develop 
a scalable method for controlled non-linear system identification:
using RFF approximation of HSE-PSR together with
a refinement procedure to enhance the model after a two-stage regression initialization. We have demonstrated promising results for the proposed method in terms of predictive performance.
As future work, we would like to use this framework for further tasks such as imitation learning and reinforcement learning.


 

\paragraph{Acknowledgements}
The authors gratefully acknowledge support from ONR (grant number N000141512365), DARPA (grant number FA87501720152)
and NSF EAGER (grant number IIS1450543).
The authors would like to thank Wen Sun and Yuxiang Wang for the helpful discussions.

\bibliographystyle{aaai}
\bibliography{controlled}

\appendix
\onecolumn
\renewcommand\thesection{\Alph{section}}
\numberwithin{equation}{section}
\numberwithin{theorem}{section}
\section{RFF-PSR Learning Algorithm}
For ease of exposition, we assume that RFF features are computed prior to PCA. 
In our implementation, we compute the RFF features on the fly while performing PCA to reduce the required memory footprint.
Here we use $A \star B$ to denote the Khatri-Rao product of two matrices (columnwise Kronecker product).

\begin{algorithm}[h]
\begin{algorithmic}
\STATE \textbf{Input:} Matrices $\Phi^h, \Phi^o, \Phi^a$  of history, observation and action features (each column corresponds to a time step). Matrices $\Psi^o, \Psi^a, {\Psi^o}', {\Psi^a}'$ of test observations, test actions, shifted test observations and shifted test actions.
\STATE \textbf{Output:} S2 regression weights $\hat{W}_\fstat$ and $\hat{W}_o$.
\STATE \textbf{Subroutines:} 
\STATE $\textsc{Svd}(X,p)$, returns the tuple $(U,U^\top X)$, where $U$ consists of top $p$ singular vectors of $X$. 
\STATE \quad  
\STATE \{Feature projection using PCA\}
\STATE $U^h, \Phi^h \leftarrow \textsc{Svd}(\Phi^h, p)$; 
\STATE $U^o, \Phi^o \leftarrow \textsc{Svd}(\Phi^o, p)$; $U^a, \Phi^a \leftarrow \textsc{Svd}(\Phi^a, p)$; 
\STATE $U^o_\psi, \Psi^o \leftarrow \textsc{Svd}(\Psi^o, p)$; $U^a_\psi, \Psi^a \leftarrow \textsc{Svd}(\Psi^a, p)$;
\STATE $U^o_\xi, \Xi^o \leftarrow \textsc{Svd}(({U^o_\psi}^\top {\Psi^o}') \star \Phi^o, p)$;
\STATE $U^a_\xi, \Xi^a \leftarrow \textsc{Svd}(\Phi^a \star ({U^a_\psi}^\top {\Psi^a}'), p)$;
\STATE $U^{oo}, \Phi^{oo} \leftarrow \textsc{Svd}(\Phi^o \star \Phi^o, p)$

\STATE \quad
\STATE \{S1 Regression and State Projection\}
\STATE Estimate $\bar{Q}_t$, $\bar{P}^\xi_t$, $\bar{P}^o_t$ for each time $t$
using the one of the S1 methods in \ref{sec:joints1rff}.
\STATE Reshape $\bar{Q}_t$, $\bar{P}_t$ as column vectors for each $t$ and then stack
the resulting vectors in matrices $\mathbf{Q}$, $\mathbf{P}^\xi$ and $\mathbf{P}^o$. 
\STATE $U^q, \mathbf{Q} \leftarrow \textsc{Svd}(\mathbf{Q}, p)$
\STATE \{S2 Regression\}
\STATE $\hat{W}_\xi \leftarrow \arg\min_{W \in \mathbb{R}^{p^2 \times p}} \|\mathbf{P}^\xi - W \mathbf{Q}\|^2 + \lambda_2 \|W\|_F^2$
\STATE $\hat{W}_o \leftarrow \arg\min_{W \in \mathbb{R}^{p^2 \times p}} \|\mathbf{P}^o - W \mathbf{Q}\|^2 + \lambda_2 \|W\|_F^2$
\end{algorithmic}
\caption{Learning Predictive State Representation with Random Fourier Features (\textsc{Learn-RFF-PSR})}
\label{alg:rffpsr}
\end{algorithm}

\section{Examples of Predictive State Controlled Models}
Here we discuss IO-HMM and Kalman filter with inputs, showing that they are instances of PSCMs.
We do this for each model by defining the predictive state, showing that it satisfies the condition $P_t = W Q_t$ and describing an S1 regression method.

\subsection{IO-HMM}
Let $T$ be the transition tensor such that
$T \times_s s_t \times_a a_t = \E[s_{t+1} | a_t, s_t]$ 
and $O$ be the observation tensor such that
$O \times_s s_t \times_a a_t = \E[o_t | a_t, s_t]$.

Define $O^k$ to be the extended observation tensor where
$O_k \times_s s_t \times_a a_{t:t+k-1} = \E[o_{t:t+k-1} | a_{t:t+k-1}, s_t]$

As a shortcut, we will denote by $T_{ij}$ the product $T \times_s e_i \times_a e_j$.

For $k=1$, we have $O^1 = O$.

For $k>1$ we can think of $a_{t:t+k-1}$ as the outer product $a_t \otimes a_{t+1:t+k}$. So we can define $O^k$ such that
\begin{align}
O^k \times_s e_i \times_{a} (e_j \otimes e_l)
= \vec(O_{ij} \otimes (O^{k-1} \times_a e_l \times_s T_{ij}))
\label{eq:ok}
\end{align}

In words, starting from state $e_i$ and applying an action $e_j$ followed 
by a sequence of $k-1$ actions denoted by indicator $e_l$. 
The expected indicator of the next $k$ observations is the outer product 
of expected observation $o_t$ (given by $O_{ij}$) with the expected indicator 
of observations $o_{t+1:t+k-1}$ as predicted by $O^{k-1}$. Note that the two expectations being multiplied are conditionally independent given the state $e_i$ and the action sequence. 

Given the tensor $O^k$ the predictive states $Q_t$ and $P_t$ are defined to be
\begin{align*}
Q_t = O^k \times_s s_t \\
P_t = O^{k+1} \times_s s_t \\
\end{align*}
Now to show that \eqref{eq:form} holds, let $\tilde{O}^k$ be a reshaping of $O^k$ into a matrix such that 
\begin{align*}
\vec(Q_t) = \tilde{O}^k s_t
\end{align*}
It follows that 
\begin{align*}
P_t = O^{k+1} \times_s s_t = O^{k+1} \times_s ((\tilde{O}^k)^+ \vec(Q_t)),
\end{align*}
which is linear in $Q_t$.

\subsubsection{S1 Regression}

Let $s_t = s(h^\infty_t)$ be the belief state at time $t$.
Note that $s_t$ is a deterministic function of the entire history.

Under a fixed policy assumption, an indicator vector of the joint observation and action assignment is an unbiased estimate of the joint probability 
table $\mathbb{P}[\stat^a_t, \fstat^a_t \mid h^\infty_t]$. An S1 regression model can be used to learn the mapping $\pstat_t \mapsto \mathbb{P}[\stat^a_t, \fstat^a_t \mid \pstat_t]$. It is then easy to estimate the conditional probability table $\bar{Q}_t$ from the joint probability table $\mathbb{P}[\stat^a_t, \fstat^a_t \mid \pstat_t]$.

We can also use the conditional S1 approach.
By exploiting the fact that 
$\stat^o_t$ is an unbiased estimate of a single column of $Q_t$ corresponding to $\stat^a_t$. We can use \eqref{eq:perror}
to learn a function $f : h_t \mapsto \bar{Q}_t$ that best matches the training examples. 

\subsection{Kalman Filter with inputs}
The Kalman filter is given by
\begin{align*}
x_t & = A x_{t-1} + B u_t + \epsilon_t \\
o_t & = C x_t + \nu_t
\end{align*}

Given a belief state $s_t \equiv \E[x_{t-1} | h^\infty_t]$ we can write the predictive state as
\begin{align*}
\E[o_{t:t+k-1} \mid s_{t}, a_{t:t+k-1} ] = \Gamma_k s_t + U_k a_{t:t+k-1},
\end{align*}
where
\begin{align*}
\Gamma_k & = \left( 
\begin{array}{c} CA \\ CA^2 \\ \vdots \\ CA^k\end{array} \right) \\
U_k & = \left( 
\begin{array}{cccccc} B  & \mathbf{0} & \dots &  & & \mathbf{0}\\ AB & B & \mathbf{0} & \dots & & \mathbf{0} \\ A^2B & AB & B & \mathbf{0} & \dots & \mathbf{0} \\ & & \vdots \\ A^{k-1}B & & \dots & & AB & B\end{array} \right) \\
\end{align*}

The extended predictive state have similar form with $\Gamma_k$ and $U_k$ replaced with $\Gamma_{k+1}$ and $U_{k+1}$. Since $U$ is fixed, keeping track of the state amounts to keeping track of
$Q_t \equiv \Gamma_k s_t$. It follows that
\begin{align*}
P_t = \Gamma_{k+1} s_t = \Gamma_{k+1} \Gamma_{k}^+ Q_t = W Q_t
\end{align*}

If $h_t$ is a linear projection of $h^\infty_t$ (e.g. stacking of a finite window of observations and actions), it can also be shown \cite{vanoverschee:96} that 
\begin{align*}
\E[Q_t | h_t] = \tilde{\Gamma}_k h_t,
\end{align*}
for some matrix $\tilde{\Gamma}_k$.

\subsubsection{S1 Regression}
Let ${\cal F}$ be the set of functions that take the form
\begin{align*}
f(\pstat)\stat^a_t = \Gamma \pstat_t + B \stat^a_t
\end{align*}

The oblique projection method \cite{vanoverschee:96} uses linear regression to estimate $\Gamma$ and $B$ (essentially solving \eqref{eq:perror}). 
Having a fixed $B$, the conditional operator is determined by $\Gamma h_t$ through an affine transformation.
Therefore we can use $\bar{Q}_t = \Gamma h_t$.

\section{Theoretical Analysis}
Let ${\cal H} = \{h_i\}_{i=1}^N$ be a set of histories generated from an i.i.d distribution. \footnote{The i.i.d property is achieved if we can restart the system or if the data collection policy 
induces an ergodic process with a stationary distribution. In the latter case, we assume the examples
are sufficiently  spaced in time to that allow the process to mix.
However, in practice, we use all examples as this makes the error only smaller.
}
We use $Q(\pstat)$ to denote $\E[Q | \pstat]$.

The main theorem in \cite{hefny:15} bounds parameter estimation error in terms of S1 regression error.
This implies that we need to analyze the properties of S1 regression to prove Theorem \ref{thm:main}. We will look at multiple scenarios where in each scenario
we develop sufficient exploration conditions and provide an S1 error bound for these conditions.

\begin{definition}[Sufficient history set]
Consider a PSCM that satisfies 
\begin{align*}
P_t = \Wsys (Q_t)
\end{align*}
A set of histories ${\cal H} = \{h_i\}_{i=1}^M$ is called a \textbf{sufficient history set} if it is sufficient to estimate $\Wsys$
using $\E[{Q}_t|\pstat_t = h]$ and $\E[{P}_t|\pstat_t = h]$ for each $h \in {\cal H}$.
\end{definition}

Note that $\Wsys$ may not be unique, we care about estimating $\Wsys Q$ for any valid $Q$. 
From the above definition, it follows that a data collection policy provides sufficient exploration
if it allows for estimating $\E[{Q}|\pstat_t = h]$ and $\E[{P}|\pstat_t = h]$ for a sufficient history set with increasing accuracy.   

\subsection{Case 1: Discrete Observations and Actions}
Consider a discrete system where 
${\cal H}$, ${\cal A}$, ${\cal A^+}$, ${\cal O}$, ${\cal O^+}$ 
are the set of all possible histories, future action sequences, extended future action sequences,
future observation sequences and extended future observation sequences respectively. 

\begin{theorem}
Assume a discrete system where the data collection policy induces an i.i.d distribution
over histories. If the policy generates each possible extended future action sequence starting from each possible history
$M$ times, then it generates an S2 training dataset of size $N = M |{\cal H}| |{\cal A^+}|$ with S1 error bound
$\efs = \sqrt{\frac{|{\cal H}| |{\cal A^+}| |{\cal O^+}|}{2M} \log \left(\frac{2 |{\cal H}| |{\cal A^+}| |{\cal O^+}|}{\delta} \right) }$
\end{theorem}

\begin{proof}
The proof follows immediately from Heoffding's inequality which bounds the error in estimating 
the probability of an event by averaging. 

Note that we need to estimate $|{\cal H}||{\cal A}||{\cal O}|$ probabilities to estimate $Q$
and $|{\cal H}||{\cal A^+}||{\cal O^+}|$ probabilities to estimate $P$.
Therefore we divide $\delta$ by $2 |{\cal H}||{\cal A^+}||{\cal O^+}|$ to correct for multiple probability estimates.
\end{proof}

\begin{remark}
Assume the system to be 1-observable, where the history and future are of length 1.
Then a consistent estimate of $Q$ and $P$ can be obtained by a consistent estimate of the joint probability table
$P(o_{t-1:t+1},a_{t-1:t+1})$.
\end{remark}

\subsection{Case 2: Continuous System}

\begin{definition}[Range and span of a policy]
Let $\pi$ be a data collection policy with a stationary distribution. 
For a random vector $X_t = f(h^\infty_t,o_{t:\infty},a_{t:\infty})$, the \textbf{range of $\pi$ on $X$} 
is the support of the stationary distribution of $X_t$ induced by the policy $\pi$
(i.e. the set of all possible values of $X_t$ that can be generated by the stationary distribution).

The \textbf{span of $\pi$ on $X$} is the subspace spanned by the range of $\pi$ on $X$.

\end{definition}
When referring to the policy range or span, we may omit the variable name when it is clear in the context.

\begin{condition}[Action span for joint S1]
Let $\pi$ be data collection policy and let $\cal H$
be the range of $\pi$ on histories.
The action span condition for joint S1 is defined as the requirement to satisfy the following:
\begin{enumerate}
\item ${\cal H}$ is a sufficient history set.
\item For any $\pstat \in {\cal H}$, the conditional covariance $\Sigma_{\stat^a|\pstat}$ is full rank.
\end{enumerate}
\label{thm:cond_joint}
\end{condition}

\begin{condition}[Action span for conditional S1]
Let $\pi$ be data collection policy and let $\cal H$
be the range of $\pi$ on histories.
The action span condition for conditional S1 is defined as the requirement to satisfy the following:
\begin{enumerate}
\item ${\cal H}$ is a sufficient history set.
\item For any $\psi^h \in {\cal H}$ and any future action feature vector $\psi^a$,
the quantity $(\psi^h \otimes \psi^a)$ is in the policy span.
\end{enumerate}
\label{thm:cond_cond}
\end{condition}

\begin{remark}
Condition \ref{thm:cond_joint} implies Condition \ref{thm:cond_cond}.
\end{remark}

\begin{assumption}[Bounded features]
We assume that $\| \psi^h \| < c_h$ for all $h \in {\cal H}$.
Also, we assume that $\| \stat^o \| \leq c_O$ and  $\| \stat^a \| \leq c_A$
for any valid future observation sequence and action sequence respectively.
\label{thm:bounded}
\end{assumption}

\begin{theorem}
Let $\pi$ be a data collection policy and let $\cal H$
be the range of $\pi$ on histories. If Assumption \ref{thm:bounded} and Condition \ref{thm:cond_cond}
are satisfied and conditional S1 regression is used with a liner model as the correct model,
then $\pi$ provides sufficient exploration and, for all $h \in {\cal H}$ 
and any $\delta \in (0,1)$ such that $N > \frac{c^2 \log(2d_h d_A / \delta)}{\lambda_{\min}(\Sigma_{\stat^h \otimes \stat^a})}$,
the following holds with probability at least $1-\delta$
\begin{align*}
\| \hat{Q}(\pstat) - Q(\pstat) \| \leq 
c_h \left( \sqrt{\frac{\lambda_{\max} (\Sigma_{\stat^o})}{\lambda_{\min} (\Sigma_{\stat^h \otimes \stat^a})}}
\left( \frac{ \sqrt{\lambda_{\min}(\Sigma_{\stat^h \otimes \stat^a})} \Delta_1 + \lambda}{\lambda_{\min}(\Sigma_{\stat^h \otimes \stat^a})(1 - \Delta_3) + \lambda}  \right)
+ \frac{\Delta_2}{\lambda_{\min}(\Sigma_{\stat^h \otimes \stat^a})(1 - \Delta_3) + \lambda} \right)
,
\end{align*}
where
\begin{align*}
\Delta_1 & = 2 c_h c_A \sqrt{\frac{\log(2 d_h d_A / \delta)}{N}} + \frac{2 \log(2 d_h d_A / \delta)}{3N} \left( \frac{c_h^2 c_A^2}{\sqrt{\lambda_{\min} (\Sigma_{\stat^h \otimes \stat^a})}} + c_h c_A \right) \\
\Delta_2 & = 2 c_O c_h c_A \sqrt{\frac{\log((d_O + d_h d_A) / \delta)}{N}} + \frac{4 c_O c_h c_A \log((d_O + d_h d_A) / \delta)}{3N} \\
\Delta_3 & = \frac{c_h^2 c_A^2 \log (2d_h d_A / \delta)}{\lambda_{\min} (\Sigma_{\stat^h \otimes \stat^a}) N}
\end{align*}
\label{thm:cont_cond}
\end{theorem}

In the following section we provide a proof sketch for the asymptotic form in Theorem \ref{thm:main}
for joint S1.

\begin{remark}[Conditioning]
It is known that linear regression converges faster if the problem is well-conditioned.
In the two stage regression we need the good conditioning of both stages-- that is,
\begin{itemize}
\item The set of training histories result in a problem $\bar{P}_t = W \bar{Q}_t$ that is well conditioned (S2 conditioning).
\item The S1 regression problem is well conditioned.
\end{itemize}
\end{remark}
The second requirement ensures that we converge fast to good estimates of $\bar{Q}_t$ and $\bar{P}_t$.
Designing exploration policies that result in well conditioned two stage regression problems is an interesting direction for future work.

\section{Proofs of theorems}
In this section we provide proofs for Theorem \ref{thm:cont_cond}. 
The asymptotic statement in Theorem \ref{thm:main} follows directly from the main theorem in \citep{hefny:15}.
We also provide a proof sketch for the joint S1 case.

The proof strategy is as follows: 
First, we use matrix concentration bounds to analyze the effect of using estimated covariance matrices. 
Then, we analyze the effect of error in covariance matrix on regression weights.
By combining the results of both analyses, we prove the desired theorems.

\begin{lemma}[Matrix Chernoff Inequality~\citep{tropp:15}]
Consider a finite sequence $\{S_k\}$ of independent, random, Hermitian matrices with common dimension $d$.
Assume that\begin{align*}
0 \leq \lambda_{\min}(S_k) \quad \text{and} \quad  \lambda_{\max}(S_k) \leq L \quad \text{for each index } k.
\end{align*}
Introduce the random matrix
\begin{align*}
Z = \sum_k S_k
\end{align*}
Define
\begin{align*}
\mu_{\min} & \equiv \lambda_{\min}(\E[Z])
\end{align*}
Then, for any $\epsilon \in [0,1)$
\begin{align*}
\Pr(\lambda_{\min}(Z) \leq (1-\epsilon) \mu_{\min}) \leq d \left[ \frac{e^{-\epsilon}}{(1-\epsilon)^{1-\epsilon}} \right]^{\mu_{\min}/L}
\leq 2d e^{-\epsilon \mu_{\min}/L}
\end{align*}
\label{thm:chernoff}
\end{lemma}

\begin{corollary}[Minimum eigenvalue of empirical covariance]
Let $X$ be a random variable of dimensionality $d$ such that $\|X\| < c$.
Let $\{x_k\}_{k=1}^N$ be $N$ i.i.d samples of the distribution of $X$.

Define
\begin{align*}
\Sigma_{X} \equiv \E[XX^\top] \text{\quad and\quad} \hat{\Sigma}_X = \frac{1}{N} \sum_{k=1}^N x_k x_k^\top 
\end{align*}
For any $\delta \in (0,1)$ such that $N > \frac{c^2 \log(2d / \delta)}{\lambda_{\min}(\Sigma_X)}$ the following holds with probability at least $1-\delta$
\begin{align*}
\lambda_{\min}(\hat{\Sigma}_X) \geq \left(1 - \frac{c^2 \log(2d / \delta)}{\lambda_{\min}(\Sigma_X) N}\right) \lambda_{\min}(\Sigma_X)
\end{align*}
\label{thm:mineig}
\end{corollary}

\begin{proof}
Define $S_k = \frac{1}{N} x_k x_k^\top$. It follows that $\lambda_{\max}(S_k) \leq L = c^2/N$ and $\mu_{\min} = \lambda_{\min}(\Sigma_X)$.
Define
\begin{align*}
\delta \equiv 2d e^{-\epsilon N \lambda_{\min}(\Sigma_X)/c^2},
\end{align*} 
which implies that
\begin{align*}
\epsilon = \frac{c^2 \log(2d / \delta)}{\lambda_{\min}(\Sigma_X) N}
\end{align*}
It follows from Lemma \ref{thm:chernoff} that $\Pr(\lambda_{\min}(\hat{\Sigma}_X) \leq (1-\epsilon) \mu_{\min}) \leq \delta$

\end{proof}

\begin{lemma}[Matrix Bernstein Inequality~\citep{tropp:15}]
Consider a finite sequence $\{S_k\}$ of independent, random matrices with common dimensions $d_1 \times d_2$. Assume that
\begin{align*}
\E[S_k] = 0 \text{ and } \|S_k\| \leq L \quad \text{for each index } k
\end{align*}
Introduce the random matrix
\begin{align*}
Z = \sum_k S_k
\end{align*}
Let $v(Z)$ be the matrix variance statistic of the sum:
\begin{align*}
v(Z) = \max \{ \|\E(ZZ^\top), \E(Z^\top Z)\| \}
\end{align*}
Then
\begin{align*}
\Pr(\|Z\| \geq t) \leq (d_1 + d_2) \exp \left( \frac{-t^2/2}{v(Z)+Lt/3} \right)
\end{align*}
\label{thm:matrix_bern}
\end{lemma}

\begin{corollary}[Error in empirical cross-covariance]
With probability at least $1 - \delta$
\begin{align*}
\|\hat{\Sigma}_{YX} - \Sigma_{YX} \| \leq \sqrt{\frac{2 \log((d_X + d_Y) / \delta) v}{N}} + \frac{2 \log((d_X + d_Y) / \delta) L}{3N},
\end{align*}
where 
\begin{align*}
L & = c_y c_x + \| \Sigma_{YX} \| \leq 2 c_y c_x \\
v & = \max(c_y^2 \| \Sigma_X \|, c_x^2 \| \Sigma_Y\|) + \| \Sigma_{YX} \| ^2 \leq 2 c_y^2 c_x^2
\end{align*}
\label{thm:matrix_bern_err}
\end{corollary}

\begin{proof}
Define $S_k = y_k x_k^\top - \Sigma_{YX}$, it follows that
\begin{align*}
\E [S_k] & = 0 \\
\|S_k\| & = \| y_k x_k^\top - \Sigma_{YX} \| \leq \|y_k\| \|x_k\| + \| \Sigma_{YX} \| \leq c_y c_x + \| \Sigma_{YX} \|\\
\| \E[ZZ^\top] \| & = \left\| \sum_{i,j} (\E [y_i x_i^\top x_j y_j^\top] - \Sigma_{YX}\Sigma_{XY}) \right\| \\
& = \left\| \sum_i (\E[\|x_i\|^2 y_i y_i^\top] - \Sigma_{YX}\Sigma_{XY}) + \sum_{i,j\neq i} (\E[y_i x_i^\top]\E[x_j y_j^\top] - \Sigma_{YX}\Sigma_{XY}) \right\| \\
& \leq N (c_x^2 \| \Sigma_Y \| + \|\Sigma_{YX}\|^2) \\
\| \E[Z^\top Z] \| & \leq N (c_y^2 \| \Sigma_X \| + \|\Sigma_{YX}\|^2) 
\end{align*}

Applying Lemma \ref{thm:matrix_bern} we get
\begin{align*}
\delta = \Pr(\|Z\| \geq Nt) \leq (d_X + d_Y) \exp \left( \frac{-N t^2/2}{v+Lt/3} \right)
\end{align*}
and hence
\begin{align*}
t^2 - \frac{2 \log((d_X + d_Y) / \delta) Lt}{3N} - \frac{2 \log((d_X + d_Y) / \delta) v}{N} \leq 0
\end{align*}
This quadratic inequality implies 
\begin{align*}
t \leq \frac{\log((d_X + d_Y) / \delta) L}{3N} + \sqrt{\frac{\log^2((d_X + d_Y) / \delta) L^2}{9N^2} + \frac{2 \log((d_X + d_Y) / \delta) v}{N}}
\end{align*}
Using the fact that $\sqrt{a^2+b^2} \leq |a|+|b|$ we get
\begin{align*}
t \leq \frac{2 \log((d_X + d_Y) / \delta) L}{3N} + \sqrt{\frac{2 \log((d_X + d_Y) / \delta) v}{N}}
\end{align*}
\end{proof}

\begin{corollary}[Normalized error in empirical covariance]
With probability at least $1 - \delta$
\begin{align*}
\|\Sigma_{X}^{-1/2}(\hat{\Sigma}_{X} - \Sigma_{X}) \| \leq 2c \sqrt{\frac{2 \log(2d / \delta)}{N}} + \frac{2 \log(2d / \delta) L}{3N},
\end{align*}
where 
\begin{align*}
L & = \frac{c^2}{\sqrt{\lambda_{\min} (\Sigma_{X})}} + c \\
\end{align*}
\label{thm:matrix_bern_err_white}
\end{corollary}

\begin{proof}
Define $S_k = \Sigma_{X}^{-1/2} x_k x_k^\top - \Sigma_{X}^{1/2}$, it follows that
\begin{align*}
\E [S_k] & = 0 \\
\|S_k\| & \leq \| \Sigma_{X}^{-1/2}\| \|x_k\|^2 + \| \Sigma_{X}^{1/2} \| \leq \frac{c^2}{\sqrt{\lambda_{\min} (\Sigma_{X})}} + c  \\
\| \E[Z^\top Z] \| = \| \E[ZZ^\top] \| & = \left\| \sum_{i,j} (\Sigma_{X}^{-1/2} \E [x_i x_i^\top x_j x_j^\top] \Sigma_{X}^{-1/2} - \Sigma_{X}) \right\| \\
& = \left\| \sum_i (\E[\|x_i\|^2 \Sigma_{X}^{-1/2} x_i x_i^\top \Sigma_{X}^{-1/2}] - \Sigma_{X}) + \sum_{i,j\neq i} (\Sigma_{X}^{-1/2}\E[x_i x_i^\top]\E[x_j x_j^\top]\Sigma_{X}^{-1/2} - \Sigma_{X}) \right\| \\
& \leq N (c_x^2 + \|\Sigma_{X}\|^2) \leq 2 N c^2
\end{align*}

Applying Lemma \ref{thm:matrix_bern} we get
\begin{align*}
\delta = \Pr(\|Z\| \geq Nt) \leq 2 d \exp \left( \frac{-N t^2/2}{2c^2+Lt/3} \right)
\end{align*}
and similar to the proof of Corollary \ref{thm:matrix_bern_err}, we can show that
\begin{align*}
t \leq \frac{2 \log(2d / \delta) L}{3N} + 2c \sqrt{\frac{\log(2d / \delta)}{N}}
\end{align*}
\end{proof}

\begin{lemma}
Let $\hat{\Sigma}_{YX} = \Sigma_{YX} + \Delta_{YX}$ and $\hat{\Sigma}_X = \Sigma_{X} + \Delta_{X}$
where $\E[\Delta_{YX}]$ and $\E[\Delta_{YX}]$ are not necessarily zero
and $\hat{\Sigma}_X$ is symmetric positive semidefinite.
Define $W = \Sigma_{YX} \Sigma_{X}^{-1}$
and $\hat{W} = \hat{\Sigma}_{YX} (\hat{\Sigma}_{X} + \lambda)^{-1}$. It follows that
\begin{align*}
\| \hat{W} - W \| \leq 
\sqrt{\frac{\lambda_{\max} (\Sigma_Y)}{\lambda_{\min} (\Sigma_X)}}
\left( \frac{ \sqrt{\lambda_{\min}(\Sigma_X)} \| \Sigma_{X}^{-1/2} \Delta_X \| + \lambda}{\lambda_{\min}(\hat{\Sigma}_X) + \lambda}  \right)
+ \frac{\| \Delta_{YX} \|}{\lambda_{\min}(\hat{\Sigma}_X) + \lambda}
\end{align*}
\label{thm:reg_psd}
\end{lemma}

\begin{proof}
\begin{align*}
\hat{W} - ً W & = \Sigma_{YX} \left((\Sigma_{X} + \Delta_X + \lambda I)^{-1} - \Sigma_{X}^{-1} \right)
+ \Delta_{YX} (\Sigma_{X} + \Delta_X + \lambda I)^{-1} = T_1 + T_2\\
\end{align*}
It follows that
\begin{align*}
\| T_2 \| & \leq \frac{\| \Delta_{YX} \|}{\lambda_{\min}(\hat{\Sigma}_X) + \lambda}
\end{align*}
As for $T_1$, using the matrix inverse Lemma $B^{-1} - A^{-1} = B^{-1}(A-B)A^{-1}$ and the fact that $\Sigma_{YX} = \Sigma_Y^{1/2} V \Sigma_X^{1/2}$,
where $V$ is a correlation matrix satisfying $\|V\| \leq 1$ we get
\begin{align*}
T_1 & = -\Sigma_{YX} \Sigma_{X}^{-1}(\Delta_X + \lambda I)(\Sigma_X + \Delta_X + \lambda I)^{-1}\\
& = -\Sigma_{Y}^{1/2} V \Sigma_{X}^{-1/2}(\Delta_X + \lambda I)(\Sigma_X + \Delta_X + \lambda I)^{-1},
\end{align*}
and hence 
\begin{align*}
\| T_1 \| & \leq \sqrt{\lambda_{\max} (\Sigma_Y)}
\left( \frac{\| \Sigma_{X}^{-1/2} \Delta_X \| + \lambda \| \Sigma_{X}^{-1/2} \|}{\lambda_{\min}(\hat{\Sigma}_X) + \lambda}  \right) \\
& = \sqrt{\frac{\lambda_{\max} (\Sigma_Y)}{\lambda_{\min} (\Sigma_X)}}
\left( \frac{ \sqrt{\lambda_{\min}(\Sigma_X)} \| \Sigma_{X}^{-1/2} \Delta_X \| + \lambda}{\lambda_{\min}(\hat{\Sigma}_X) + \lambda}  \right) \\
\end{align*}
\end{proof}

\begin{corollary}
Let ${x_k}_{k=1}^N$ and ${y_k}_{k=1}^N$ be i.i.d samples from two random variables $X$ and $Y$
with dimensions $d_X$ and $d_Y$ and (uncentered) covariances
$\Sigma_X$ and $\Sigma_Y$ respectively. Assume $\|X\| \leq c_x$ and $\|Y\| \leq c_y$.
Let $\hat{\Sigma}_{YX} = \frac{1}{N} \sum_{k=1}^N y_k x_k^\top $ and $\hat{\Sigma}_X = \frac{1}{N} \sum_{k=1}^N x_k x_k^\top$.
Define $W = \Sigma_{YX} \Sigma_{X}^{-1}$
and $\hat{W} = \hat{\Sigma}_{YX} (\hat{\Sigma}_{X} + \lambda)^{-1}$.

For any $\delta \in (0,1)$ such that $N > \frac{c_x^2 \log(2d_X / \delta)}{\lambda_{\min}(\Sigma_X)}$ the following holds with probability at least $1-3 \delta$
\begin{align*}
\| \hat{W} - W \| \leq 
\sqrt{\frac{\lambda_{\max} (\Sigma_Y)}{\lambda_{\min} (\Sigma_X)}}
\left( \frac{ \sqrt{\lambda_{\min}(\Sigma_X)} \Delta_1 + \lambda}{\lambda_{\min}(\Sigma_X) (1 - \Delta_3) + \lambda}  \right)
+ \frac{\Delta_2}{\lambda_{\min}(\Sigma_X) (1 - \Delta_3) + \lambda}
,
\end{align*}
where
\begin{align*}
\Delta_1 & = 2 c_x \sqrt{\frac{\log(2 d_X / \delta)}{N}} + \frac{2 \log(2 d_X / \delta)}{3N} \left( \frac{c_x^2}{\sqrt{\lambda_{\min} (\Sigma_{X})}} + c_x \right) \\
\Delta_2 & = 2 c_y c_x \sqrt{\frac{\log((d_Y + d_X) / \delta)}{N}} + \frac{4 c_y c_x \log((d_Y + d_X) / \delta)}{3N} \\
\Delta_3 & = \frac{c_x^2 \log (2d_X / \delta)}{\lambda_{\min} (\Sigma_{X}) N}
\end{align*}
\label{thm:regc_psd}
\end{corollary}
\begin{proof}
This corollary follows simply from applying Corollaries \ref{thm:mineig}, \ref{thm:matrix_bern_err} and \ref{thm:matrix_bern_err_white}
to Lemma \ref{thm:reg_psd}. The $1-3 \delta$ bound follows from union bound; since we have three probabilitic bounds each of which holds with probability $1-\delta$.
\end{proof}

\begin{lemma}
Let $\hat{\Sigma}_{YX} = \Sigma_{YX} + \Delta_{YX}$ and $\hat{\Sigma}_X = \Sigma_{X} + \Delta_{X}$
where $\E[\Delta_{YX}]$ and $\E[\Delta_{YX}]$ is not necessarily zero
and $\hat{\Sigma}_X$ is symmetric but not necessarily positive semidefinite.
Define $W = \Sigma_{YX} \Sigma_{X}^{-1}$
and $\hat{W} = \hat{\Sigma}_{YX} \hat{\Sigma}_{X} (\hat{\Sigma}_{X}^2  + \lambda)^{-1}$. It follows that
\begin{align*}
\| \hat{W} - W \| \leq 
\sqrt{\frac{\lambda_{\max} (\Sigma_Y)}{\lambda_{\min}^3 (\Sigma_X)}}
\frac{\| \Delta_X \|^2 + 2 \lambda_{\max}(\Sigma_X) \| \Delta_X \| + \lambda}{\lambda^2_{\min}(\hat{\Sigma}_X) + \lambda}
+ \frac{\| \Sigma_{YX} \| \| \Delta_X \| + \| \Delta_{YX} \| \| \Sigma_X \| + \| \Delta_{YX} \| \| \Delta_X \|}{\lambda^2_{\min}(\hat{\Sigma}_X) + \lambda}
\end{align*}
\label{thm:reg_nopsd}
\end{lemma}

\begin{proof}
\begin{align*}
\hat{W} - ً W & = (\Sigma_{YX} + \Delta_{YX}) (\Sigma_{X} + \Delta_{X}) ((\Sigma_{X} + \Delta_{X})^2 + \lambda I)^{-1} - \Sigma_{YX} \Sigma_X \Sigma_X^{-2} \\
& = \Sigma_{YX} \Sigma_X (((\Sigma_X + \Delta_X)^2+\lambda I)^{-1}-\Sigma_X^{-2}) 
+ (\Sigma_{YX}\Delta_X + \Delta_{YX}\Sigma_X + \Delta_{YX}\Delta_X)((\Sigma_{X} + \Delta_{X})^2 + \lambda I)^{-1} \\
& = T_1 + T_2
\end{align*}

Using the matrix inverse Lemma $B^{-1} - A^{-1} = B^{-1}(A-B)A^{-1}$ and the fact that $\Sigma_{YX} = \Sigma_Y^{1/2} V \Sigma_X^{1/2}$,
where $V$ is a correlation matrix satisfying $\|V\| \leq 1$ we get
\begin{align*}
T_1 & = - \Sigma_{YX}^{1/2} V \Sigma_{X}^{-3/2} (\Delta_X^2 + \Sigma_X \Delta_X + \Delta_X \Sigma_X + \lambda I) ((\Sigma_X + \Delta_X)^2 + \lambda I)^{-1} \\
\| T_1 \| & \leq \sqrt{\frac{\lambda_{\max} (\Sigma_Y)}{\lambda_{\min}^3 (\Sigma_X)}}
\frac{\| \Delta_X \|^2 + 2 \lambda_{\max}(\Sigma_X) \| \Delta_X \| + \lambda}{\lambda^2_{\min}(\hat{\Sigma}_X) + \lambda} \\
\| T_2 \| & \leq \frac{\| \Sigma_{YX} \| \| \Delta_X \| + \| \Delta_{YX} \| \| \Sigma_X \| + \| \Delta_{YX} \| \| \Delta_X \|}{\lambda^2_{\min}(\hat{\Sigma}_X) + \lambda}
\end{align*}
\end{proof}

\subsection{Proof of Theorem \ref{thm:cont_cond}}
\begin{proof}
In the linear case, we estimate a tensor $T$ with modes corresponding to $\stat^h$, $\stat^a$ and $\stat^o$ by solving the minimization problem in Section \ref{sec:conds1rff}.
Equivalently, we estimate a matrix $T_r$ of size $d_O \times d_h d_A$ where an input $\pstat \otimes \psi^a$ is mapped to an output $\E[\psi^o \mid h, \psi^a]$.
Note that 
\begin{align*}
Q(\pstat) \stat^a = T \times_h \pstat \times_A \stat^a = T_r (\pstat \otimes \stat^a)
\end{align*}
For any history $h \in {\cal H}$ and future action feature vector $\stat^a$ we have
\begin{align*}
\| \hat{Q}(\pstat) - Q(\pstat)\| & = \mathrm{argmax}_{\stat^a} \frac{\| (\hat{Q}(\pstat) - Q(\pstat)) \stat^a \|}{ \| \stat^a \| } \\
& = \mathrm{argmax}_{\stat^a} \frac{\| (\hat{T}_r - T_r) (\pstat \otimes \stat^a) \|}{\| \stat^a \|} \leq \|\hat{T}_r - T_r\| \|\psi^h\|
\end{align*}
Note that Condition \ref{thm:cond_cond} implies that $\pstat \otimes \stat^a$ will eventually be in the span of training examples.
This rules out the case where the inequality is satisfied only because $(\pstat \otimes \stat^a)$ is incorrectly in the null space of $\hat{T}_r$ and $T_r$.
 
The theorem is proven by applying Corollary \ref{thm:regc_psd} to bound $\|\hat{T}_r - T_r\|$. 
\end{proof}

\subsection{Sketch Proof for Joint S1}
Let $T_A$ be a tensor such that $\Sigma_{\stat^a \mid \pstat} = T_A \times_h \pstat$
In order to prove Theorem \ref{thm:main} for joint S1, note that
\begin{align*}
\| \hat{\Sigma}_{\stat^a|\pstat} - \Sigma_{\stat^a|\pstat} \| & \leq \| \hat{T}_A - T_A \| \| \pstat \| \\
\| \hat{\Sigma}_{\stat^o \stat^a|\pstat} - \Sigma_{\stat^o \stat^a|\pstat} \| & \leq \| \hat{T}_{OA} - T_{OA} \| \| \pstat \| \\
\end{align*}
From Lemma \ref{thm:reg_psd}, we obtain a high probability bound on $\| \hat{T}_A - T_A \|$ and $\| \hat{T}_{OA} - T_{OA} \|$.
Then we apply these bounds to Lemma \ref{thm:reg_nopsd} to obtain an error in $Q(\pstat)$.

\end{document}